\documentclass{article}

\usepackage[preprint]{neurips_2019}

\usepackage[utf8]{inputenc} 
\usepackage[T1]{fontenc}    
\usepackage{hyperref}       
\usepackage{url}            
\usepackage{booktabs}       
\usepackage{comment}       
\usepackage{amsfonts}       
\usepackage{nicefrac}       
\usepackage{microtype}      

\usepackage[pdftex]{graphicx}

\usepackage{xcolor}
\usepackage{subcaption}

\title{Compressing Representations for Embedded Deep Learning}

\author{
  Juliano S. Assine*\And Alan Godoy$\dagger$\And Eduardo Valle*
  \\
  {*}RECOD Lab. -- School of Electrical and Computer Engineering\\
  University of Campinas -- UNICAMP, São Paulo, Brazil \\
  \texttt{jsiloto,dovalle@dca.fee.unicamp.br} \\
  \\
  {$\dagger$}QuintoAndar, Inc., São Paulo, Brazil \\
  \texttt{alan.godoy@quintoandar.com.br} \\
}

\begin{document}

\maketitle

\begin{abstract}
Despite recent advances in architectures for mobile  devices, deep learning computational requirements remains prohibitive for most embedded devices. 
To address that issue, we envision sharing the computational costs of inference between local devices and the cloud, taking advantage of the compression performed by the first layers of the networks to reduce communication costs. Inference in such distributed setting would allow new applications, but requires balancing a triple trade-off between computation cost, communication bandwidth, and model accuracy. We explore that trade-off by studying the compressibility of representations at different stages of MobileNetV2, showing those results agree with theoretical intuitions about deep learning, and that an optimal splitting layer for network can be found with a simple PCA-based compression scheme.
\end{abstract}

\section{Introduction}
\label{sec:introduction}

Although desktops and servers represent the most visible aspect of computational hardware, embedded systems represent the vast majority of computational systems, not only in raw number of devices, but also in market size~\citep{embedded_market_by_2019}. Today's embedded systems comprise ``classical'' cases, like controllers for consumer electronics, vehicles, and industrial processes, but also smartphones and tablets, and the so-called Internet of Things (IoT), which is bringing computation to a myriad of everyday devices.

Implementing deep learning in embedded systems is extremely challenging, since those systems have strict limitations of computation, memory, power consumption and network access. So far, even the most efficient image classification models are still supported only by high-end mobile devices with gigabytes of RAM~\citep{ignatov2018ai}, while technological studies on IoT and constrained computation usually consider nodes with memory in the kilobyte range~\citep{bormann2014terminology}.

In this work, we will formalize the problems of designing deep learning models for IoT systems, and present one case study of distributed inference that illustrates the interplay between computing power, network bandwidth and accuracy common in those scenarios. Finally we will show how we obtained a satisfactory solution by applying intuitions brought by recent theoretical results found in the literature of deep learning.

\section{Partial Computation on Embedded Devices}
\label{sec:formalization}

We will tackle the problem of partial computation on the edge by splitting the inference performed by the model during test in two sequencial steps, first on the local device, and next on the cloud. 

Let us denote $f_i$ each of the successive layers of a deep-learning model ($f_0$ being the input, and $f_n$ the output layer).
Each layer $x_{i+1}=f_i(x_i)$ turns an input representation $x_i$ into an output representation $x_{i+1}$. The entire network $y=f(x_0)$ turns the model input $x_0$ into the final decision $y$ (where $f=f_0\circ\ldots\circ f_n$). In our context, optimizing such model consists in picking a layer $k$, $0 \leq k \leq n+1$, such that $x_k=f_L(x_0)$ is computed locally in the embedded system, $x_k$ is transmitted over the network, and $y=x_{n+1}=f_R(x_k)$ is executed remotely in the cloud (where $f_L=f_0\circ\ldots\circ f_{k-1}$ and $f_R=f_k\circ\ldots\circ f_n$). For $k=0$, everything runs on the cloud, and for $k=n+1$, everything runs locally. 

Building feasible solutions, however, involves much more than picking an architecture, attempting $k$ choices for the cut and choosing the best. Such a naïve approach would lead, most of the time, to unfeasible (or grossly suboptimal) solutions. Several additional techniques must be considered. Compressing/distilling the \textit{model}~\citep{han2016deepcompression,hinton2015distillation,sheng2018quantization} reduces its computation cost, allowing to process larger fractions of it locally. Compressing the intermediate \textit{representation}~\citep{carvalho2016deep,vogel2019guaranteed} reduces the network bandwidth needed to offload the final inference step to the cloud. In addition to distillation and compression, ``native'' mobile-friendly deep learning architectures have been proposed~\citep{sandler2018mobilenetv2, tan2019mnasnet}, but for implementation in IoT, even those networks might be too expensive. Distillation and compression affect accuracy, so applying them involves balancing a delicate compromise.

That problem can be formulated in a number of ways, applying different objectives, costs and constraints --- those include model accuracy, computational capabilities of the local and the remote devices, energy limitations of the local device, latency and bandwidth of the network, process queuing at the remote device, and (monetary) operational costs of using the network or the remote device. 

One frequent scenario in computer vision is to maximize model accuracy while minimizing operational costs under the constraint of a time deadline. To process $v$ video frames per second, we have a strict deadline of $v^{-1}$ seconds to deliver the decision. The straightforward solution would be to process it locally or transmit the video over the network and process it remotely but, today, few configurations of model/device/network allow feasible solutions for real-time computer-vision: deep learning models are way too computationally expensive to run on most embedded devices and high-bandwidth/low-latency internet is either not available or prohibitively expensive in most of the world.

Such distributed inference could have many applications, which existing approaches fail to solve adequately. Lowering computational and communication costs may have profound impacts on tasks that require continuous influx of data, but not continuous feedback to the device: activity detection, surveillance and remote sensing. Even where high-bandwidth, low-latency networks (notably 5G) are available, distributed inference can reduce the workload of cellphones, tablets, virtual reality headsets, etc., keeping heat dissipation and battery draining at acceptable levels, while still providing a fluid experience to the user.

\section{Experiments}
\label{sec:experiments}

In this paper we consider a scenario without the constraint of a deadline, but where we still want to minimize network costs. A scenario of image classification --- here, ImageNet~\citep{ILSVRC15} --- although simplified, captures many characteristics of the economic model of IoT applications, where the main compromises are spending computation (and thus energy/battery) on the local device, and spending network bandwidth (which has costs for both users and hosts). We simplify the optimization by considering that the cloud servers are so much more powerful than the user device, that once a transaction goes to the cloud, we pay fixed costs for it, independent of $k$ and $|x_k|$. 

Optimizing the model for this scenario involves a very complex optimization:

\begin{itemize}
    \item Choosing and optimizing the architecture. That is probably the most thorny choice, since, for the moment, model architecture design is mostly an \textit{ad hoc} endeavor;
    \item Model distribution across local and remote nodes, given by the parameter $k$, as explained above; 
    \item Model distillation, or representation compression, or both.
\end{itemize}

\begin{figure}[h!]
    \centering
    \includegraphics[width=\textwidth]{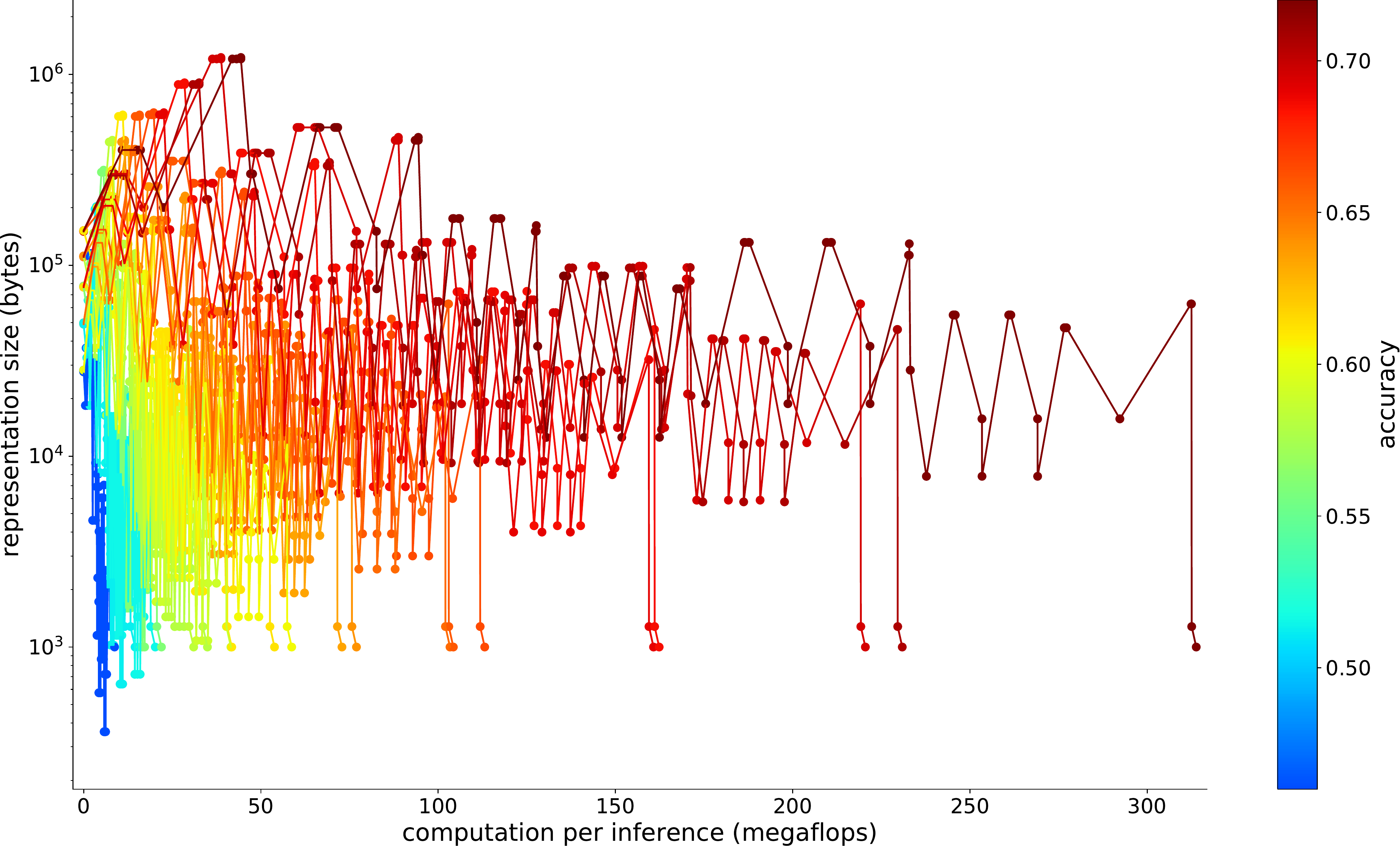}
   \caption{\textbf{The triple trade-off of computation spent on the mobile device, network cost, and task accuracy}. We run our experiments on MobileNetV2 for the ImageNet task. Each series represents one configuration ($\alpha$, \textit{input size}) for the model, and each dot represents one layer of the model. Picking a layer to slice the model (parameter $k$ in Section~\ref{sec:formalization}) is key for optimizing the problem.}
   \label{fig:compromises}
\end{figure}

For our experiments we picked MobileNetV2~\citep{sandler2018mobilenetv2}, an architecture already optimized for small devices. Fig.~\ref{fig:compromises} illustrates the different compromises obtainable by varying MobileNetV2 ``built-in'' parameters $\alpha$ and \textit{input size} --- each series/line in the plot representing a model variation. Each dot represents a layer on the model, and thus a different choice for the parameter $k$ as explained above. Each choice represents a compromise between accuracy, network costs, and computation/energy expenditure on the user device --- the former given by the model choice and the latter two given by $k$.

\begin{figure}[t!]
    \includegraphics[width=0.98\textwidth]{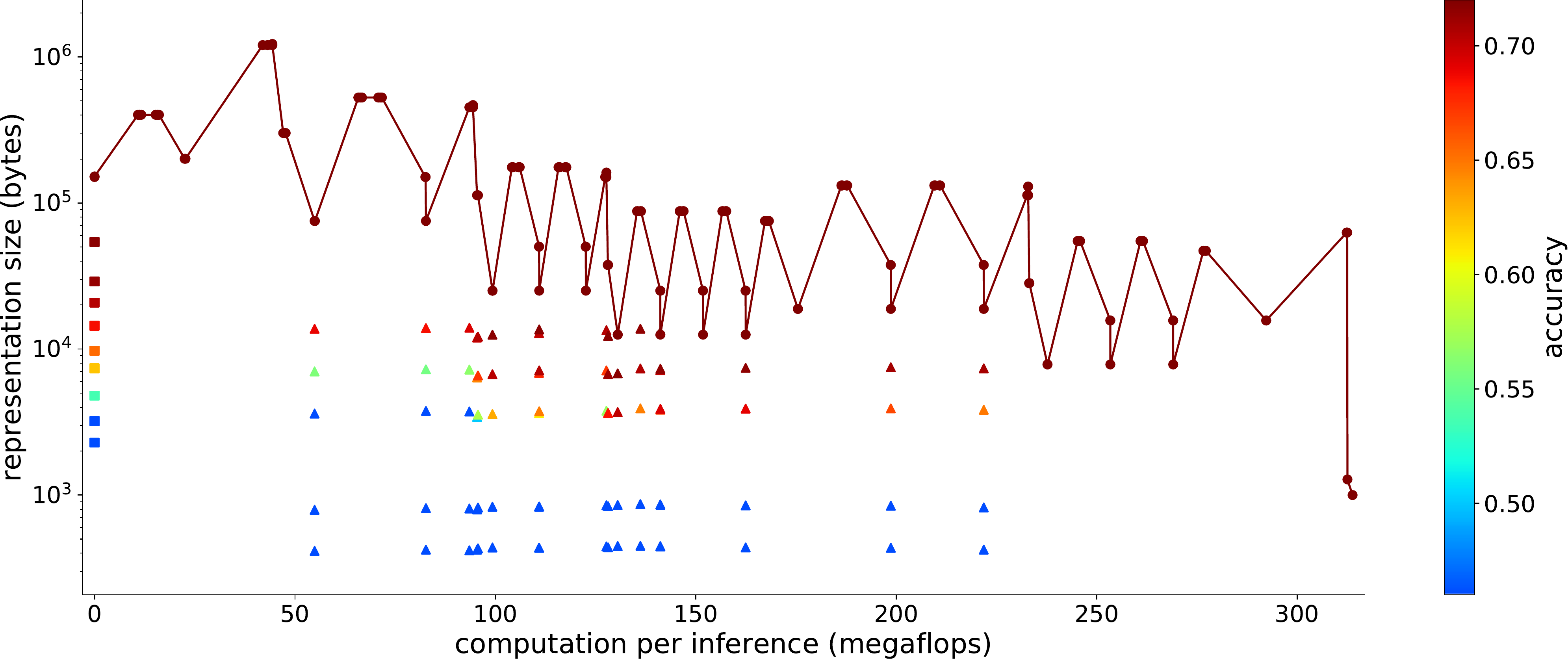} \\ \\
    \includegraphics[width=0.98\textwidth]{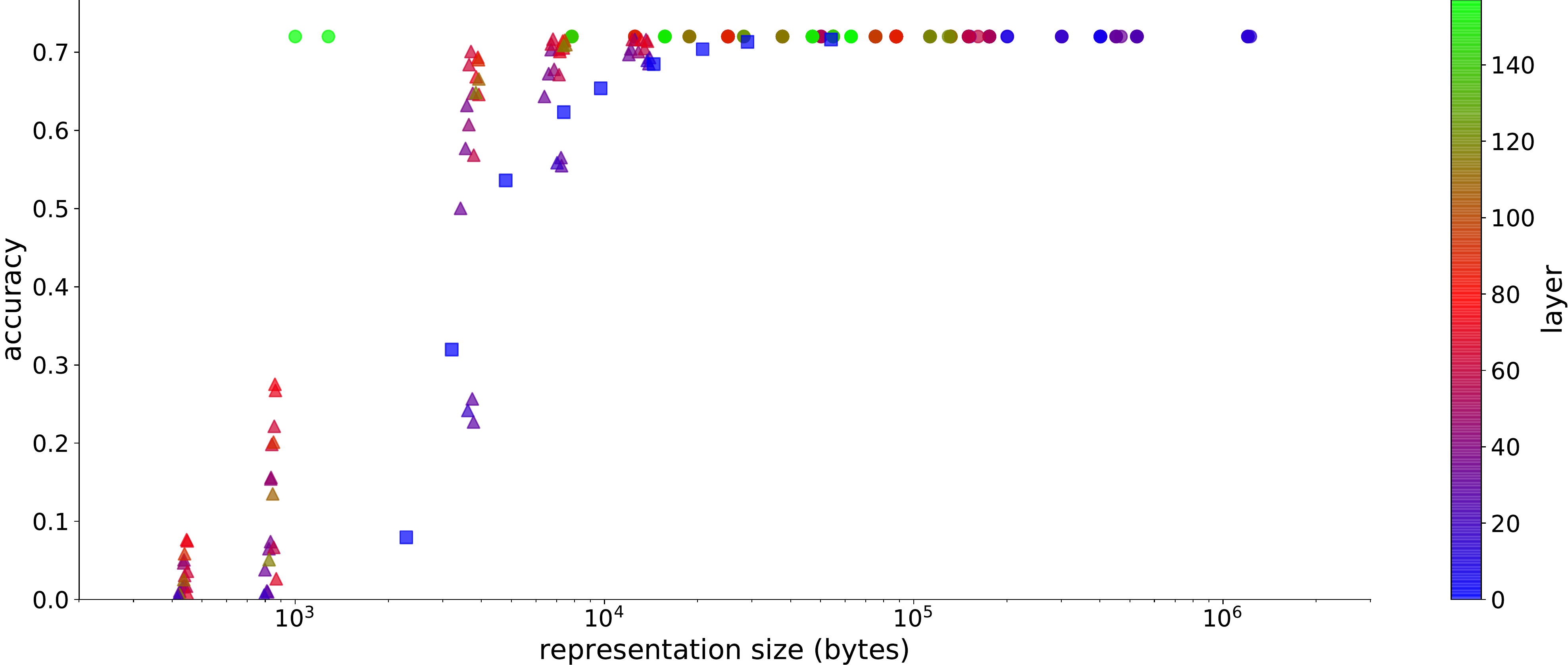}    \caption{\textbf{Applying compression to the representations}. We pick the MobileNetV2 model with best accuracy ($\alpha=1.0$, \textit{input size}$=224$) and optimize it by applying a PCA-based compression to the output of different layers (triangles). As predicted by the Information Bottleneck and the linearization of deep representations, there is a sweet spot around the middle of the network, much more interesting than simply compressing the input with JPEG (squares).}
    \label{fig:compression}
\end{figure}

As Fig.~\ref{fig:compression} shows, those compromises change radically when representation compression is employed. In that experiment, we applied a straightforward compression scheme using blocked principal component analysis (PCA) (in which we divide the data into blocks and compress them independently) followed by quantization and Huffman coding. The choices of the previous paragraph still apply, but now we can compress the output $x_{k}$ of $f_{k-1}$ before sending it to the cloud, which allows greatly reducing the bandwidth costs. Compression also introduces an increase in computational costs at the client, but in our tests we found those overheads to be negligible in comparison to the costs of deep learning, and we will ignore them on this motivating example. As shown in the plots, we find that applying the compression of intermediate layers allows using the best version of MobileNetV2 at a small fraction of the bandwidth costs of transmitting JPEG compressed images to the cloud.

Insights from information theory help guide the design and understanding of these experiments~(Figs.~\ref{fig:compromises} and~\ref{fig:compression}). For example, if we consider that each representation is sampled from a random variable $x_i\sim X_i$, and that the model output $y=x_{n+1}\sim\hat Y$, is drawn from an estimator of the true labels $Y$, the well-known data-processing inequality establishes that mutual information $I(x_0; x_i) \leq I(x_0; x_j)$ for $i>j$, and thus, although the \textit{data structure} of the representation of later layers may, in some circumstances, take more storage, the information they have about the input may never grow.

The Information Bottleneck \citep{tishby2015deep} provides additional insights, establishing that well-trained networks tend to discard most information about their input except that which is useful for the task --- suggesting, for example, that the output of the last layers of the network is already highly compressed.

Furthermore, we expect those deep representations to behave somewhat linearly. \citep{wu2016learning} shows that deep learning tends to create representations that ``make sense'' in linear spaces, and the design of MobileNetV2 architecture also explicitly encourages that by avoiding non-linear activations.

Our results agreed surprisingly well with the theoretical intuitions: we found a definite sweet spot by picking $k$ in the middle of the network  (Fig.~~\ref{fig:compression}). We found that earlier layers do not compress well with PCA and require complex algorithms (like the JPEG compression we used on the input), while later layers are already very small and any significant (relative) compression tends to penalize accuracies too much. As also predicted by the Information Bottleneck, the effective information about the input only decreases as the layers progress, even if the actual data structure increases --- as their compressibility vs. accuracy shows.

We hope that the example above shows that information-theoretic insights and compression of deep representations can be fruitful in the progress of embedded deep learning, first, by enabling applications otherwise infeasible, and second, by providing a rich playground where theoretical results may have significant practical impacts.

\subsubsection*{Acknowledgments}
This work is partially funded by CNPq grant 424958/2016-3. J. S. Assine is funded by the Coordenação de Aperfeiçoamento de Pessoal de Nível Superior - Brasil (CAPES) - Finance Code 001. E. Valle is partially funded by CNPq grant PQ-2 311905/2017-0 and FAPESP grant 2019/05018-1. The RECOD Lab receives additional funds from FAPESP, CNPq, and CAPES. We gratefully acknowledge NVIDIA for the donation of GPU hardware. 
The authors J. S. Assine and A. Godoy would also like to thanks the CPQD research foundation where this research was partially conducted.

\medskip

\small
\bibliographystyle{apalike}
\bibliography{bibliography.bib}

\end{document}